\documentclass[final,1p,times,twocolumn,authoryear]{elsarticle}

\usepackage{amssymb}
\usepackage{lipsum}
\usepackage{amsmath}
\usepackage{float}

\usepackage{url}

\usepackage{breakurl}
\usepackage[breaklinks]{hyperref}

\journal{Information Processing \& Management}

\begin{document}

\begin{frontmatter}



\title{Training Gradient Boosted Decision Trees on Tabular Data Containing Label Noise for Classification Tasks}


\author[Debeka]{Anita Eisenbürger}
\author[Debeka]{Daniel Otten}
\author[Debeka,HSKo]{Anselm Hudde}
\author[UK,TUoS]{Frank Hopfgartner}

\affiliation[Debeka]{organization={Debeka},
             city={Koblenz},
             country={Germany}
            }

\affiliation[UK]{organization={Universität Koblenz, Institute for Web Science \& Technologies},
             city={Koblenz},
             country={Germany}
            }            

\affiliation[TUoS]{organization={University of Sheffield},
             city={Sheffield},
             country={United Kingdom}
            }            

\affiliation[HSKo]{organization={Hochschule Koblenz, Department of Math and Technology},
             city={Remagen},
             country={Germany}
            }

\begin{abstract}
Label noise, which refers to the mislabeling of instances in a dataset, can significantly impair classifier performance, increase model complexity, and affect feature selection. While most research has concentrated on deep neural networks for image and text data, this study explores the impact of label noise on gradient-boosted decision trees (GBDTs), the leading algorithm for tabular data. This research fills a gap by examining the robustness of GBDTs to label noise, focusing on adapting two noise detection methods from deep learning for use with GBDTs and introducing a new detection method called Gradients.
Additionally, we extend a method initially designed for GBDTs to incorporate relabeling. By using diverse datasets such as Covertype and Breast Cancer, we systematically introduce varying levels of label noise and evaluate the effectiveness of early stopping and noise detection methods in maintaining model performance. Our noise detection methods achieve state-of-the-art results, with a noise detection accuracy above 99\% on the Adult dataset across all noise levels. This work enhances the understanding of label noise in GBDTs and provides a foundation for future research in noise detection and correction methods.
\end{abstract}


\begin{keyword}
Label noise \sep Gradient-boosted decision trees \sep Data quality \sep Data cleansing

\end{keyword}

\end{frontmatter}

\section{Introduction}
\label{introduction}
Accurate predictive modeling is essential in machine learning applications such as healthcare diagnostics and financial risk assessment. For tabular data, gradient boosted decision trees (GBDTs) are among the top-performing algorithms, often surpassing deep learning models in both accuracy and efficiency \citep{tree-better-dnn}. However, one major challenge in deploying GBDTs in real-world settings is the presence of label noise—incorrect labels assigned to training instances—which can degrade model performance significantly.

Label noise differs from feature noise, as it directly impacts the learning process by corrupting the target variable rather than the feature set. Mislabeled data, often introduced through human error, is costly to eliminate and typically harder to detect than feature noise \citep{ln-survey}. This issue is particularly pressing for GBDTs on tabular data, where research on label noise detection and correction is limited, with most label noise research focusing on training deep learning models on image or text data \citep{dividemix, nn-dropout}. 

This work aims to bridge this gap by exploring label-noise robust methods for GBDTs in tabular data settings. Specifically, we seek to (1) provide a comprehensive overview of state-of-the-art label noise handling techniques, (2) enhance these techniques by developing a robust GBDT model for noisy labels, and (3) evaluate the model’s performance relative to existing classifiers. 

Following the introduction, Section \ref{sec:related} provides an overview of label noise research with a focus on deep learning and GBDTs. Section \ref{sec:prelim} discusses the problem formulation and taxonomy of label noise, while Section \ref{sec:method} explains the methodology, including noise detection and correction techniques. Section \ref{sec:experiments} describes the experimental setup, covering noise generation, evaluation metrics, and model configuration. Results are presented in Section \ref{sec:result}, and Section \ref{sec:discussion} offers an analysis of the findings. Lastly, Section \ref{sec:conclusion} summarizes the main conclusions and outlines future research directions.

The main contributions are as follows: 
\begin{itemize}
    \item This study addresses the gap in label noise research specifically focused on gradient-boosted decision trees (GBDTs) for tabular data.
    \item Two noise detection methods designed for deep neural networks were adapted for GBDTs, and a novel detection method, Gradients, was introduced.
    \item Experiments reveal that GBDTs are naturally robust to label noise, particularly symmetric noise, with early stopping improving performance.
    \item State-of-the-art noise detection accuracy is achieved with AUM and LRT, both of which can reliably estimate the level of noise in a dataset.
\end{itemize}

\section{Related Work} \label{sec:related}
This section reviews primary approaches for handling label noise and emphasizes recent developments in deep learning and gradient-boosted decision trees (GBDTs).

\subsection{Approaches to Label Noise}
Label noise handling typically falls into three main categories: robust models, noise-tolerant learning algorithms, and data cleansing methods \citep{ln-survey}. Label noise robust models, such as bagging or variations of Adaboost, are based on algorithms that aren't gravely affected by label noise in the first place; however, their effectiveness is often limited to simpler noise patterns \citep{bagging-better, robust-problem}. Noise-tolerant algorithms, including probabilistic and frequentist approaches, either model label noise or regularize models to reduce their sensitivity to noisy instances \citep{bayes-prior, freq-mixture}. Lastly, data cleansing methods, which identify and remove or relabel noisy instances, are widely used for their simplicity but risk excluding valuable data or introducing further errors when relabeling \citep{anomaly-measure, coteach}.

\subsection{Deep Learning and Label Noise}
In deep learning, common practices include adding robust architectures that model the noise transition matrix, regularization, improved loss functions and data cleansing, where instances with a likely true label are chosen to update the network \citep{deep-survey}. To mitigate the accumulation of errors when further training is based on the network's own predictions, some researchers have investigated the use of multiple networks, where each network selects low-loss instances for the other to use in backpropagation \citep{deVos2023StochasticCF}. Robust DivideMix \newline\citep{badlabel} operates by having each network discard the labels of high-loss instances and then provides the resulting partially-labeled data to the other network for semi-supervised training. Additionally, NRAT \citep{Chen2024}, a robust adversarial training method, combines a robust loss function and enhanced regularization to improve adversarial robustness under label noise.

Deep learning achieves state-of-the-art performance on image and text data, but the methods often suffer from slow convergence, high computational cost due to an increased number of hyperparameters or large models due to deep architectures \citep{deep-survey}. Techniques specifically designed for tabular data remain scarce, as DNN-based methods are typically optimized for unstructured data types.

\subsection{GBDTs and Label Noise in Tabular Data}
GBDTs are widely regarded as the leading algorithm for tabular data, offering significant efficiency and performance advantages over DNNs in structured datasets \citep{tree-better-dnn}. Despite this, research on how GBDTs handle label noise remains relatively sparse. \cite{training-dynamics} introduced an approach to detect and remove mislabeled instances in GBDTs by calculating the confidence, correctness, and variability of predictions during training, enhancing dataset quality. \cite{adapting-influence} adapted three established methods for estimating the influence of training samples on predictions, which can help identify correct, mislabeled or problematic samples. \cite{johnson2024label} examined loss design techniques from deep learning to improve the robustness of decision trees, while \cite{zhu2024tree} leveraged isolation forests to detect noisy instances, which were subsequently relabeled using a semi-supervised learning algorithm.




\section{Preliminaries}\label{sec:prelim}
Label noise is commonly categorized into three types: Noisy Completely at Random (NCAR), Noisy at Random (NAR), and Noisy Not at Random (NNAR) \citep{ln-survey,deep-survey}.

In NCAR noise, labels are flipped randomly, independent of instance features or class, leading to symmetric misclassification. This can be modeled with a noise transition matrix $S \in [0,1]^{c \times c}$, where $S_{ij} := p(\tilde{y} = j | y = i)$ represents the probability of the true label $i$ being misclassified as label $j$, and $c$ denotes the number of classes. For NCAR noise with rate $\tau \in [0,1]$, the transition matrix is defined as $S_{ii} = 1 - \tau$ for correct labels and $S_{ij} = \frac{\tau}{c-1}$ for $i \neq j$.

NAR noise, also called class-dependent noise, introduces asymmetry in the label flipping process. In this case, the mislabeling probability depends on the class but remains independent of instance features. Thus, we have $S_{ii} = 1 - \tau$, while $S_{ij} \neq S_{ik}$ for certain $k \neq i$, allowing for pairwise noise where similar classes are more frequently confused (e.g., a "bird" class being misclassified as "plane" rather than "dog").

NNAR noise is the most complex type, with mislabeling probabilities influenced by both instance features and true class. Here, the misclassification probability is defined as $p(\tilde{y} = j | y = i, x)$, where $x$ denotes instance features. In practice, NNAR noise simulates real-world scenarios where specific regions in feature space are more prone to errors, such as when instances are visually or statistically similar to those in other classes.

\begin{figure}[H]
\caption{Noise transition matrices for a dataset with four classes on no noise, 20\% symmetric and 20\% pair noise, respectively. }
\centering
\includegraphics[width=0.69\textwidth]{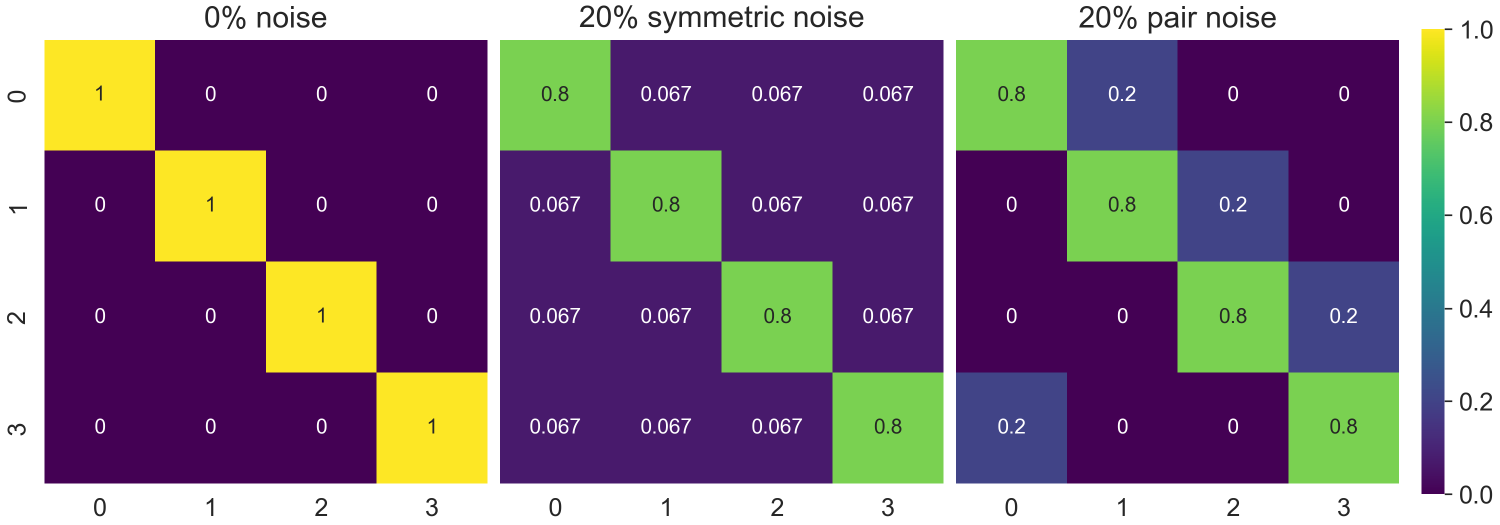}
\label{fig:noise_tm_example}
\end{figure}

Figure \ref{fig:noise_tm_example} shows noise transition matrices for a dataset with four classes under no noise, 20\% symmetric (NCAR), and 20\% pair (NAR) noise. Given the complexity of generating NNAR noise, and for comparability with other studies, this work focuses on NCAR and NAR noise, hereafter referred to as "symmetric" and "pair" noise, respectively.

\section{Methodology} \label{sec:method}
This section presents the noise detection and correction methods implemented and adapted for this work. 

\subsection{Noise Detection Methods}
Four methods were implemented for detecting noisy instances: \textit{LRT-Correction}, \textit{AUM Ranking}, \textit{Training Dynamics Statistics}, and \textit{Gradients}. \textit{LRT-Correction} and \textit{AUM Ranking} were adapted from deep neural networks (DNNs), while \textit{Training Dynamics Statistics} is based on prior GBDT research. The \textit{Gradients} method, a novel approach here, is inspired by weight clipping in Adaboost and the small-loss trick.

\subsubsection{LRT-Correction}
LRT-Correction (Likelihood Ratio Test Correction) uses a likelihood ratio to evaluate the purity of an instance's label. Let $p(i|x)$ denote the classifier’s predicted probability for class $i$, and $\hat{y} = \arg\max_i p(i|x)$ represent the predicted class. The likelihood ratio $LR(x, \tilde{y}) = \frac{p(\tilde{y}|x)}{p(\hat{y}|x)}$ is calculated between the classifier’s confidence in the noisy label $\tilde{y}$ and its prediction $\hat{y}$ \citep{adacorr}:
\begin{align}
    \tilde{y}_{new} = 
    \begin{cases}
        \hat{y}, & \text{if } LR(x, \tilde{y}) < \epsilon, \\
        \tilde{y}, & \text{otherwise}.
    \end{cases}
\end{align}

The original approach also includes an additional retroactive loss term to improve label consistency across epochs, although this term was not implemented in this study. Here, the likelihood ratio $LR(x, \tilde{y})$ is used to identify noisy labels, combined with either relabeling or removal of these instances (see Section \ref{sec:combinations}).

\subsubsection{AUM Ranking}
"Area Under the Margin" (AUM) Ranking uses margins at each epoch to classify instances as easy, hard, or mislabeled. The margin $M$ at epoch $t$ measures the difference between the assigned logit $z^t_{\tilde{y}}(x)$ and the highest logit of another class, with $\operatorname{AUM}$ representing the average margin over all epochs \citep{aum}:
\begin{align}
    M^t(x,\tilde{y}) &= z^t_{\tilde{y}}(x) - \underset{i\neq\tilde{y}}{\max}\;z_i^t(x),\\
    \operatorname{AUM}(x, \tilde{y}) &= \frac{1}{T} \sum^T_{t=1} M^t(x, \tilde{y}).
\end{align}

A positive margin indicates confidence in prediction, while a negative margin suggests uncertainty or mislabeling. Although \cite{aum} also categorize threshold instances into a separate class to better isolate mislabeled data, this work focuses on the AUM metric solely for identifying noisy instances.

\subsubsection{Training Dynamics Statistics} 
Training dynamics capture a model's behavior during training and are used in data cartography to classify instances as "easy-to-learn," "hard-to-learn," or "ambiguous." This approach uses the predicted probability $p(i|x)$ and predictions $\hat{y}$ across estimators $t$. Confidence $\mu(x)$ is the average probability for the true label $\tilde{y}$, variability $\sigma(x)$ measures fluctuation in true class probabilities, and correctness $\gamma(x)$ represents the percentage of correct classifications \citep{dataset-cartography}:
\begin{align}
    \mu(x) &= \frac{1}{T} \sum^T_{t=1} p^t(\tilde{y}|x), \\
    \sigma(x) &= \sqrt{\frac{\sum_{t=1}^T (p^t(\tilde{y}|x) - \mu(x))^2 }{T}}, \\
    \gamma(x) &= \frac{1}{T} \sum^T_{t=1} [\hat{y}=\tilde{y}],
\end{align}
where $T$ is the number of trees.

\cite{training-dynamics} applied dataset cartography to GBDTs, using confidence and correctness to filter problematic instances through thresholding or re-weighting. Here, we extend the thresholding method to also relabel identified instances, denoted as \textit{ConfCorr} in Section \ref{sec:experiments}.

\subsubsection{Gradients} 
This method combines the "small-loss trick" from deep learning with the growing gradients of noisy instances observed in boosting algorithms \citep{ada-remove}. The small-loss trick assumes that low-loss samples are likely clean, updating the network with these instances \citep{coteach, mentornet, dividemix}. Although gradient boosting does not assign instance weights, the per-instance gradients serve a similar purpose.

\begin{figure}[H]
\caption{Maximum absolute gradients of noisy and clean instances per epoch (Covertype). Noisy instances exhibit significantly larger gradients. }
\centering
\includegraphics[width=0.3\textwidth]{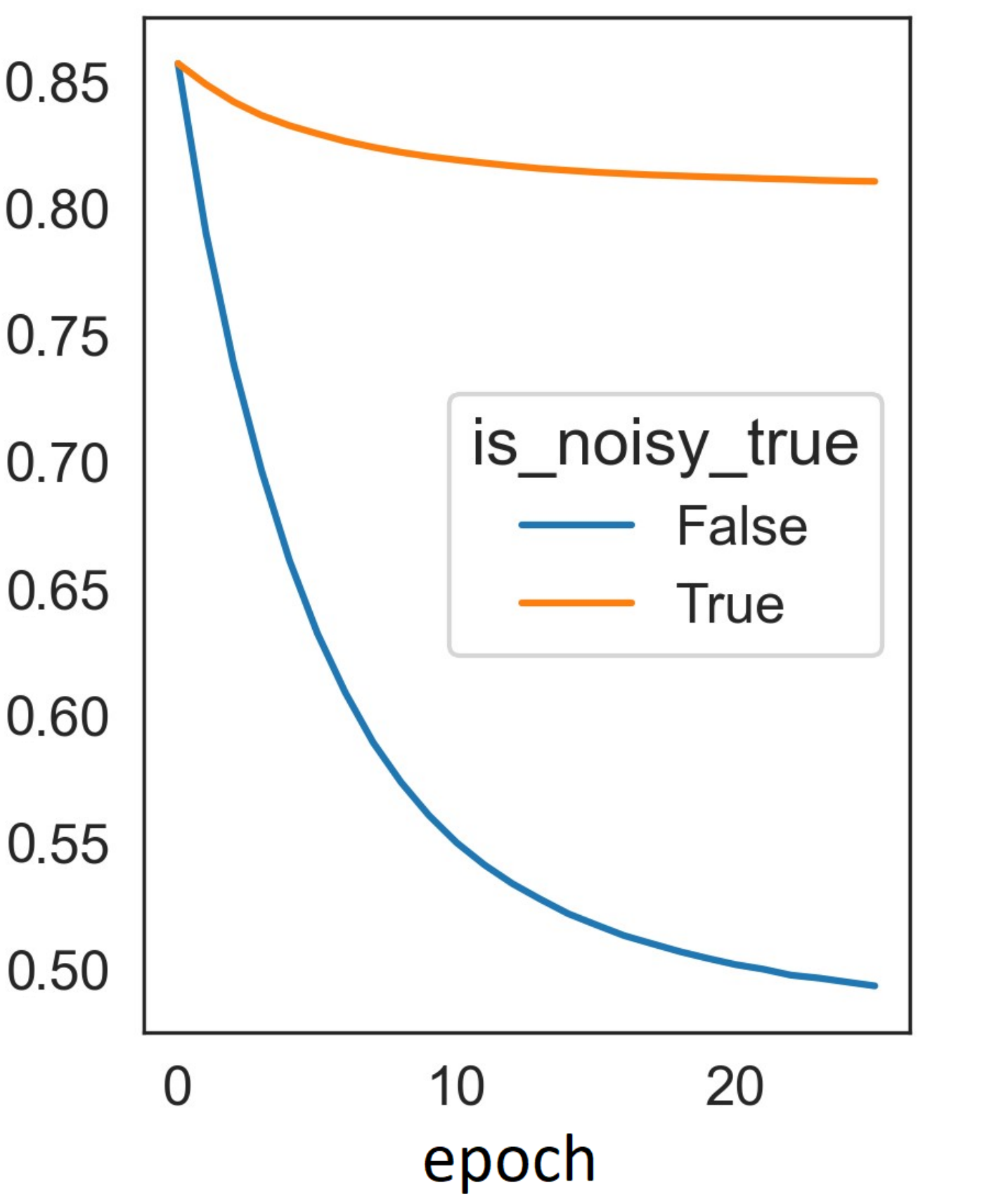}
\label{fig:gradients_epoch}
\end{figure}

An experiment on the Covertype dataset shows that noisy instances display larger gradients than clean instances, with noisy gradients averaging 1.5 times higher by epoch ten (see Figure \ref{fig:gradients_epoch}).

Following \cite{dividemix}, a two-component Gaussian Mixture Model (GMM) is fitted to the loss distribution over the final epochs, with the component of smaller mean indicating clean probabilities. The largest absolute gradient per instance represents its noisiness in this model.

\subsection{Noise Correction} \label{sec:combinations}
This work focuses on two noise correction methods: removal and relabeling. While other methods like instance reweighting and loss correction exist, they are beyond this study's scope.

Each noise detection method was paired with either removal or relabeling, yielding eight combinations. 
For relabeling, instances were reassigned to the class with the highest probability averaged over a history window and could only be reassigned once. To preserve sufficient data for training, removal was limited to at most 80\% of the entire dataset.


\section{Experiments}\label{sec:experiments}
This section outlines the datasets, noise injection process, evaluation metrics, model configuration, and experimental settings used in this study. 

\subsection{Datasets}
This study evaluates model performance on four public datasets frequently used in label noise and GBDT research: Covertype \footnote{\url{https://archive.ics.uci.edu/dataset/31/covertype}, accessed 30th July 2024}, Dry Bean \footnote{\url{https://archive.ics.uci.edu/dataset/602/dry+bean+dataset}, accessed 30th July 2024}, Adult \footnote{\url{https://archive.ics.uci.edu/dataset/2/adult}, accessed 30th July 2024} and Breast Cancer \footnote{\url{https://archive.ics.uci.edu/dataset/17/breast+cancer+wisconsin+diagnostic}, accessed 30th July 2024}. These datasets represent diverse classification tasks and data types, as summarized in Table \ref{tab:datasets}. Each dataset varies significantly in size and degree of class imbalance, with Covertype being the largest and most imbalanced.

\begin{table}[h!] 
\centering
\begin{tabular}{c c c c c}  
\hline
Dataset & \#Instances & \#Features & \#Classes & Types \\ 
\hline
 Covertype & 581012 & 54 & 7 & Num \\
 Dry Bean & 13611 & 16 & 7 & Num \\
 Adult & 48842 & 14 & 2 & Cat/Num \\ 
 Breast Cancer & 569 & 30 & 2 & Num \\
\hline
\end{tabular}
\caption{Characteristics of the datasets used in this study.}
\label{tab:datasets}
\end{table}

The labels are assumed to be nearly clean, given their sources: Covertype and Adult rely on reliable datasets, while Dry Bean and Breast Cancer are professionally labeled. Preprocessing included imputing missing values, standardizing numeric attributes, and one-hot encoding categorical features.

\subsection{Noise Injection}
Assuming that all datasets are sufficiently clean, a noise transition matrix for pair and symmetric noise was applied to the training labels as defined in Section \ref{sec:prelim}. The methods were evaluated on seven noise rates, $\tau \in \{0.0, 0.1, 0.2, 0.3, 0.4, 0.5, 0.6\}$, for each type of noise. In the case of pair noise, noise rates exceeding 50\% make misclassification more likely than the correct label. However, these higher rates were still included to assess their impact on the model's performance. In the final comparison (Section \ref{sec:final-results}), only noise rates ranging from 10\% to 40\% were considered.

\subsection{Evaluation Metrics}
Classification performance is evaluated using test set accuracy, precision, recall, and F1-score, with test accuracy based on true labels. Training accuracy, measured with noisy labels, is shown only in the experimental figures, not in the final comparison to other works. For noise detection, accuracy is reported because it is a widely used metric across most papers, ensuring comparability with prior research.

\subsection{Model and Hyperparameters}\label{sec:params}
For the classification task, a GBDT model was trained using the XGBoost library. The default parameters proved to be the most effective: maximum tree depth was set at 6, learning rate at 0.3, and the tree method was \texttt{auto}. 

Because the low-level boosting interface was used, early stopping had to be implemented manually with a minimum required loss improvement of 0.5 and a patience of 10 epochs. The model optimized the cross-entropy objective function and had a warm-up duration of 15 epochs. In deep learning, warm-up refers to the initial part of training where hyperparameters, especially the learning rate, are gradually increased from a lower value to the target value. This approach is employed to avoid instability or divergence of the training process, ensuring a more stable and effective convergence of the model. In the context of GBDTs, the term ``warm-up" is be used to describe training the model for a few iterations to ensure that noise detection methods have enough past predictions to calculate their metrics from. 

XGBoost's \texttt{softprob} and \texttt{logistic} objective functions were used for multiclass and binary classification, respectively. 

For both \textit{Gradients} and \textit{ConfCorr}, employing a two-component GMM to determine the optimal threshold at each iteration proved more effective in distinguishing clean from noisy instances than using a fixed or dynamic threshold. The threshold for \textit{LRT} was based on its originating paper's recommendations, setting the $\delta$ parameter to 1.0. Given that a positive margin indicates a correct prediction and a negative one an incorrect prediction, as per its original paper, the threshold for \textit{AUM} was set to 0. 

Consistent with the original \textit{ConfCorr} implementation, metrics are calculated incrementally across all epochs. The remaining noise detection methods calculate metrics over a history of five epochs.

\subsection{Experimental Settings} \label{sec:algo}
The experimental setup is divided into three stages. In the first stage, the impact of label noise on GBDTs is assessed by training the models on datasets with noisy labels and evaluating them on clean datasets. 

In the second part of the experiments, the noise detection and correction methods are applied and compared to each other. Each detection method is paired with each correction method and evaluated against a baseline model (denoted by \textit{none}), which is a GBDT trained without any noise-handling measures and serves as a control to gauge the effects of noise detection and correction by comparing them to an uncorrected model.

Finally, in the third stage, the performance of the implemented methods are compared to the work of \cite{training-dynamics}.

\section{Results} \label{sec:result}
The following sections present the results of experiments on label noise in GBDTs. The impact of label noise without correction is first examined, followed by a comparison of the implemented noise detection and correction techniques. Finally, these methods are compared to a state-of-the-art approach.

\subsection{Effect of Label Noise on Training GBDTs}
This section examines the effects of label noise on GBDTs when not correcting for label noise. To get a measure for the model's classification performance, the performance of the trained model was evaluated on a clean test set, unpolluted by noise, and is denoted as ``test" in the following figures. 

The Covertype and Dry Bean datasets, both multiclass, were chosen to enable generalization to binary classification tasks. The complex Covertype dataset served as the primary source for figures, with the Dry Bean dataset used for cases where Covertype's size proved computationally limiting. Only one representative noise type (symmetric or pair) is shown if both produced similar trends.

\begin{figure}[H]
\caption{The types of predictions the model makes during training at 10\% noise (Dry Bean). Only instances where the true label deviates from the noisy label are shown. The model predicts mostly the ground-truth true label in the beginning and gradually adapts to the noisy labels. }
\centering
\includegraphics[width=0.89\textwidth]{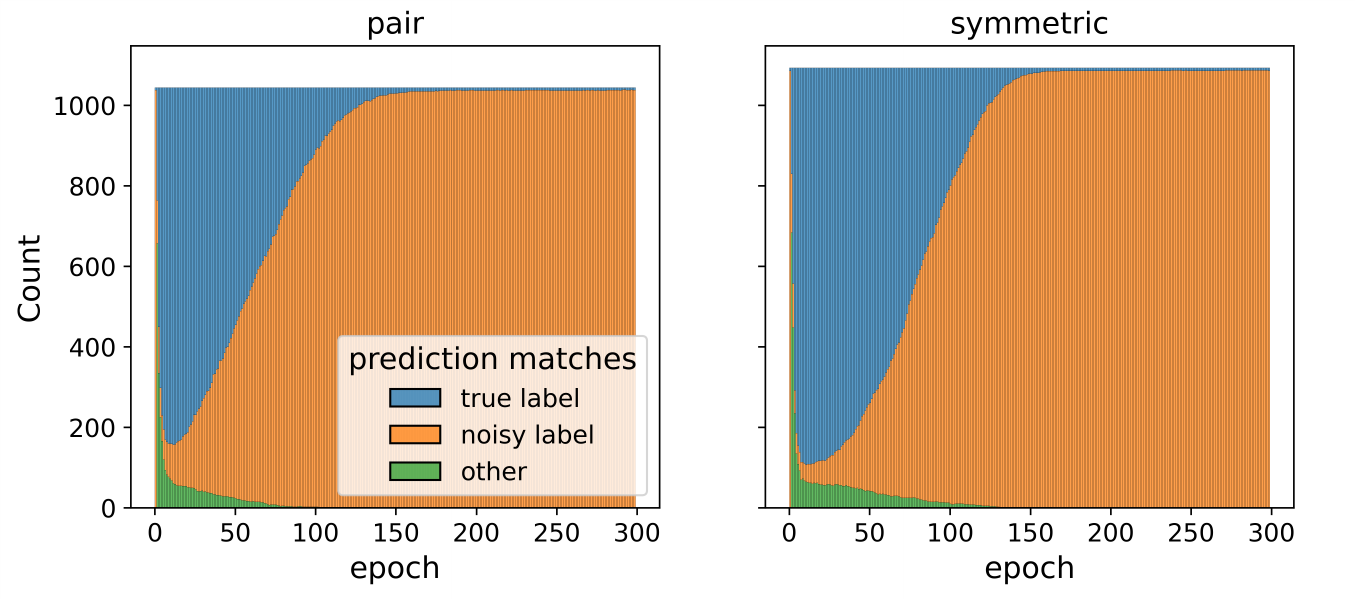}
\label{fig:pred_type_curve}
\end{figure}

To find out how naturally robust GBDTs are to label noise and what types of errors the model makes as training progresses, the model was first trained without early stopping for 100 epochs and its predictions were collected and categorized into three groups:  the prediction matches (1) the ground-truth label, (2) the noisy label or (3) another class altogether. Figure \ref{fig:pred_type_curve} shows that for both noise types, the model correctly predicts the true class most of the time at the start of training. The number of correct predictions decreases as training progresses and the number of predictions that match the noisy label grows. At about 60 epochs with 10\% pair noise and 95 epochs with symmetric noise, predictions that match with the true and noisy label are equally likely. Only few predictions match a label that is neither the noisy nor the true label in the beginning and this number decreases as training progresses. 

\begin{figure}[H]
\caption{Classification accuracy on the train and test set per epoch at 30\% noise without early stopping (Dry Bean). Test set accuracy on the clean test set decreases much slower when trained on symmetric noise, implying that GBDTs are more robust towards symmetric noise.}
\centering
\includegraphics[width=0.89\textwidth]{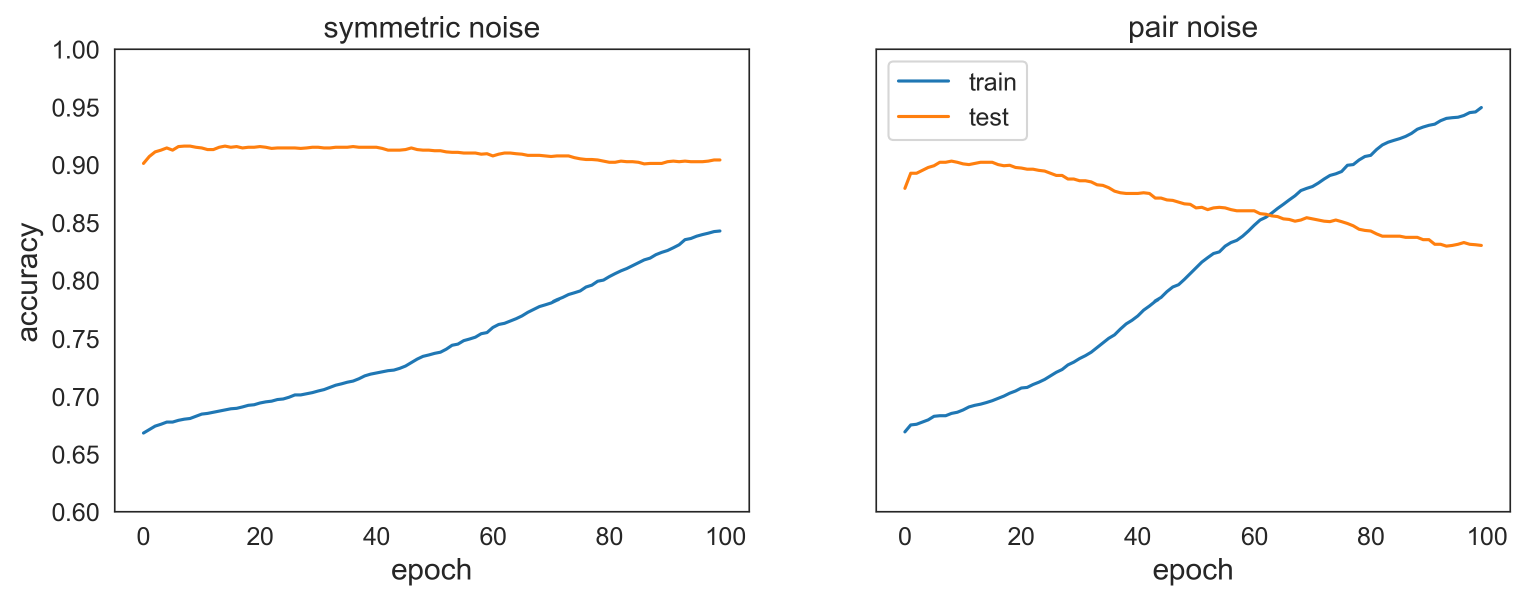}
\label{fig:train_accuracy}
\end{figure}

Furthermore, Figure \ref{fig:train_accuracy} shows the classification accuracy on the train and test set per epoch. In both plots, the accuracy on the training set increases while that on the test set decreases as training progresses. When trained on pair noise, the decrease of accuracy on the test set and increase of the accuracy on the training set is much more rapid than when trained on symmetric noise. The increase in training accuracy is due to the model gradually adapting to the noisy data, as illustrated above. And because the model learns the noisy patterns, the test set accuracy decreases as training progresses. The test set accuracy is initially higher than the train set accuracy, because the test set accuracy is evaluated on the clean labels whereas the train set accuracy is evaluated on the noisy labels. The difference between the initial test and train set accuracy is approximately equal to the noise rate, i.e. 30\%. 

From those two figures one can infer that GBDTs, despite the fact that boosting algorithms are known to be very sensitive to noise, actually mostly predict the correct label and only gradually adapt to the noisy labels. Therefore early stopping is an important tool to use for training.

\begin{figure}[H]
\caption{Training curves at 10\% and 40\% pair noise respectively (Dry Bean). The difference in performance is smaller on the test set than the train set, implying the model is also somewhat robust to pair noise. }
\centering
\includegraphics[width=0.89\textwidth]{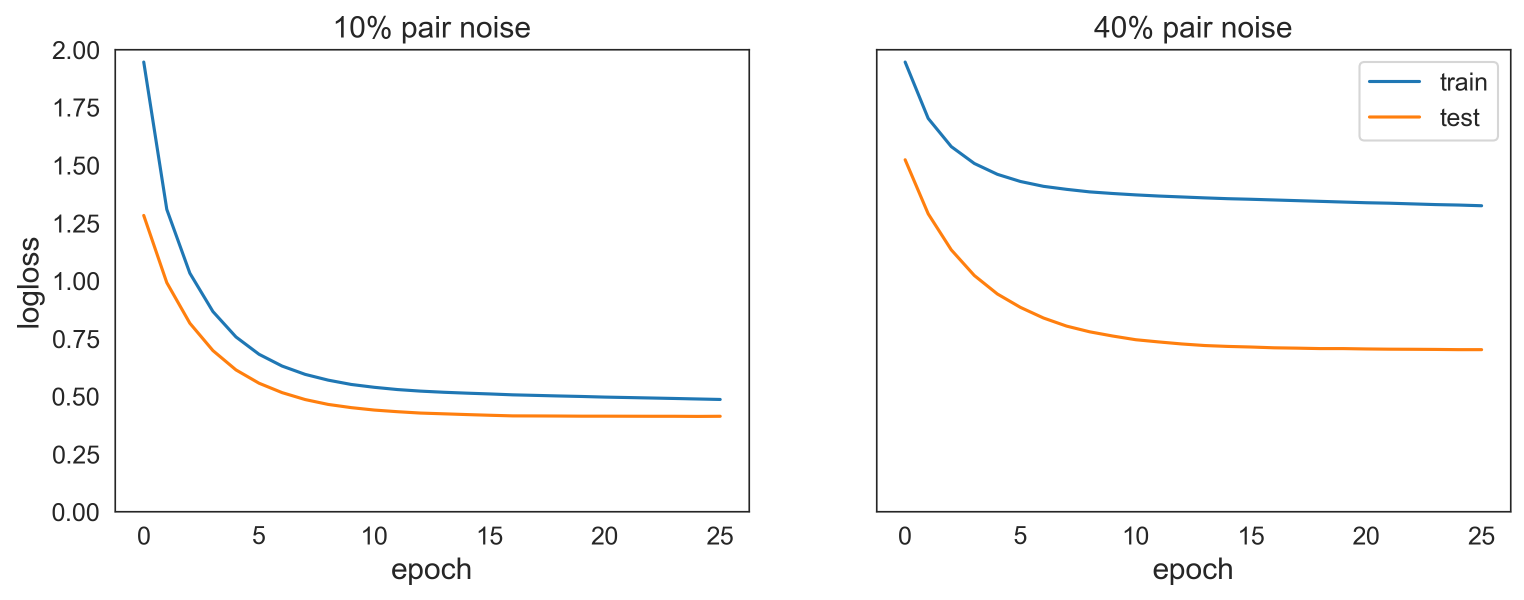}
\label{fig:train-curve-rate}
\end{figure}

Figure \ref{fig:train-curve-rate} visualizes the training curves on 10\% and 40\% pair noise, respectively. Train set performance converges at a much lower logloss with 10\% noise than with 40\% noise. Test set performance also converges at a lower logloss with 10\%  noise, but the difference in test set performance is smaller than in training set performance. The substantial difference in training set performance highlights the adverse impact of label noise on classification performance when evaluated on noisy data. Meanwhile, the smaller difference in test set performance suggests the robustness of GBDTs to label noise.

In summary, while symmetric noise appears harder for the model to learn, it results in better logloss performance on a clean test set. The training curves (Figure \ref{fig:train_accuracy}) confirm that, depending on when training is stopped, training on symmetric noise results in a higher classification accuracy on the test set than training on pair noise. 

\subsection{Noise Detection and Correction}
In this section, the performance of the implemented noise detection and correction methods is compared. 
 
\begin{figure}[H]
\caption{Distribution of the calculated metrics by the ground truth noisiness of a label (Covertype), calculated on 30\% pair noise. \textit{Gradients} and \textit{AUM} seem to separate between noisy and clean instances best. }
\centering
\includegraphics[width=0.98\textwidth]{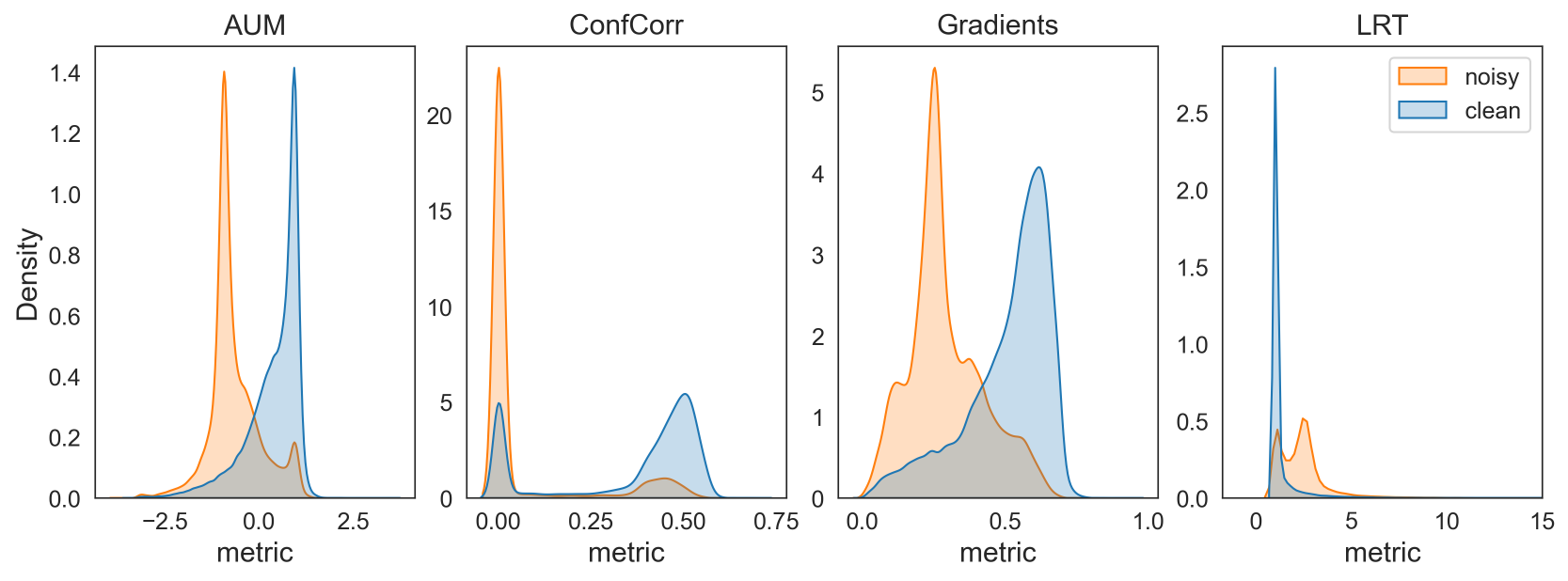}
\label{fig:crit-density}
\end{figure}

Figure \ref{fig:crit-density} visualizes the metrics calculated by each of the implemented noise detection methods, separated by whether the instance the value was calculated for was noisy or clean. For effective differentiation between noisy and clean instances, it is desirable for the two distributions to be distinct and well-separated. The opposite would mean that the values calculated for noisy and clean instances don't differ a lot and therefore the method couldn't be reliably used to identify noisy instances. 

The metric for \textit{AUM} forms two close but distinct distributions of equal height, intersecting at a value of zero. The distributions for \textit{ConfCorr} are widely separated; however, a significant number of clean instances fall within the noisy distribution, and a few noisy instances are found in the clean distribution. The \textit{Gradients} distribution also forms two distinct distributions, but they are less peaked than those of \textit{AUM}, with an intersection around 0.4. The clean distribution for \textit{LRT} is tightly clustered around the value zero, although some noisy instances also fall within this range, with most distributed over values greater than zero.

From this, one might infer that \textit{AUM} and \textit{Gradients} could be the most effective in separating clean from noisy instances, given their unimodal distributions. Their performance might be followed by \textit{ConfCorr}, whose distributions are bimodal and have more overlap but still maintain a considerable gap. Although \textit{LRT} clusters all clean instances closely around a single value, the presence of some noisy instances within this distribution suggests that \textit{LRT}'s performance might be on the lower end. 

\begin{figure}[H]
\caption{Noise detection accuracy of all detection methods at various noise rates, calculated at the first epoch after warm-up (Dry Bean). All methods achieve a noise detection accuracy of 70\% or higher on both noise types, with \textit{LRT} and \textit{AUM} achieving the highest accuracy of at least 90\%.} 
\centering
\includegraphics[width=0.79\textwidth]{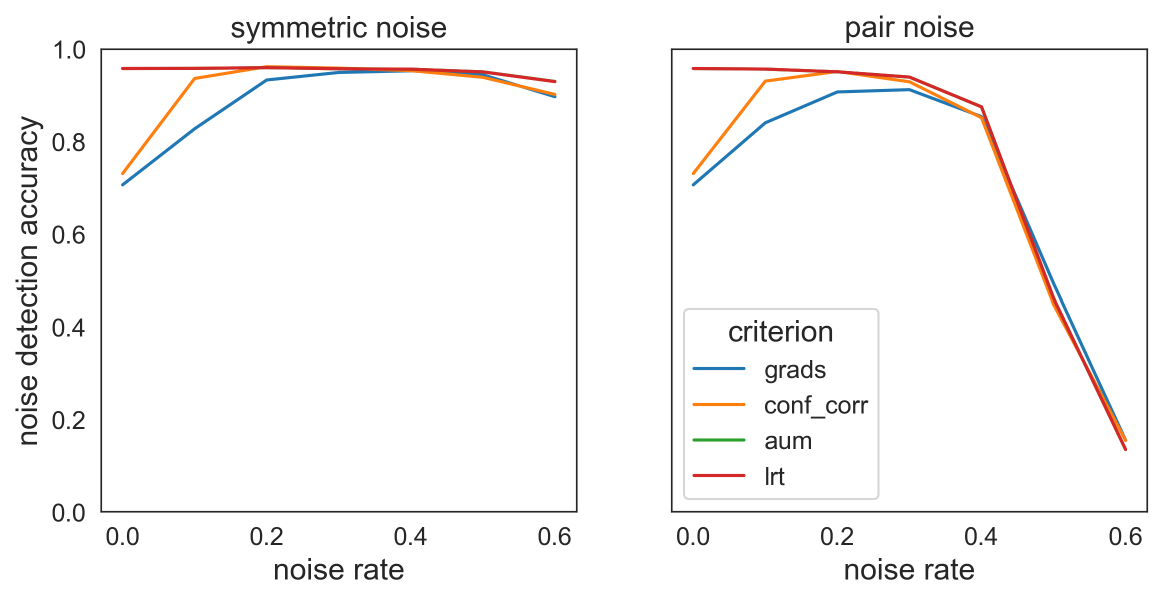}
\label{fig:noise_det_acc_all_bean}
\end{figure}

Figure \ref{fig:noise_det_acc_all_bean} plots the noise rate against noise detection accuracy. All methods achieve a noise detection accuracy of 70\% or higher on both noise types. \textit{LRT} and \textit{AUM} overlap, which is why \textit{AUM}'s line isn't visible. They both also perform the best across all noise rates on symmetric noise, always achieving an accuracy of at least 90\%. \textit{ConfCorr}'s and \textit{Gradients}' accuracy is slightly worse for low and high noise rates. Pair noise degrades the quality of the data more quickly than symmetric noise, which is also reflected in the right plot of Figure \ref{fig:noise_det_acc_all_bean}. The correctness of the label is completely random at 50\% pair noise, and likewise all methods reach a noise detection accuracy of about 50\% and decreases further afterwards. 

To summarize, \textit{LRT} and \textit{AUM} achieve the highest noise detection accuracy across all noise rates on the Dry Bean dataset, closely followed by \textit{Gradients} and \textit{ConfCorr}. 

\begin{figure}[H]
\caption{Noise detection accuracy of all detection methods at various noise rates, calculated at the first epoch after warm-up (Dry Bean). The percentage of instances marked as noisy by \textit{LRT} and \textit{AUM} is close to the actual noise rate. }
\centering
\includegraphics[width=0.79\textwidth]{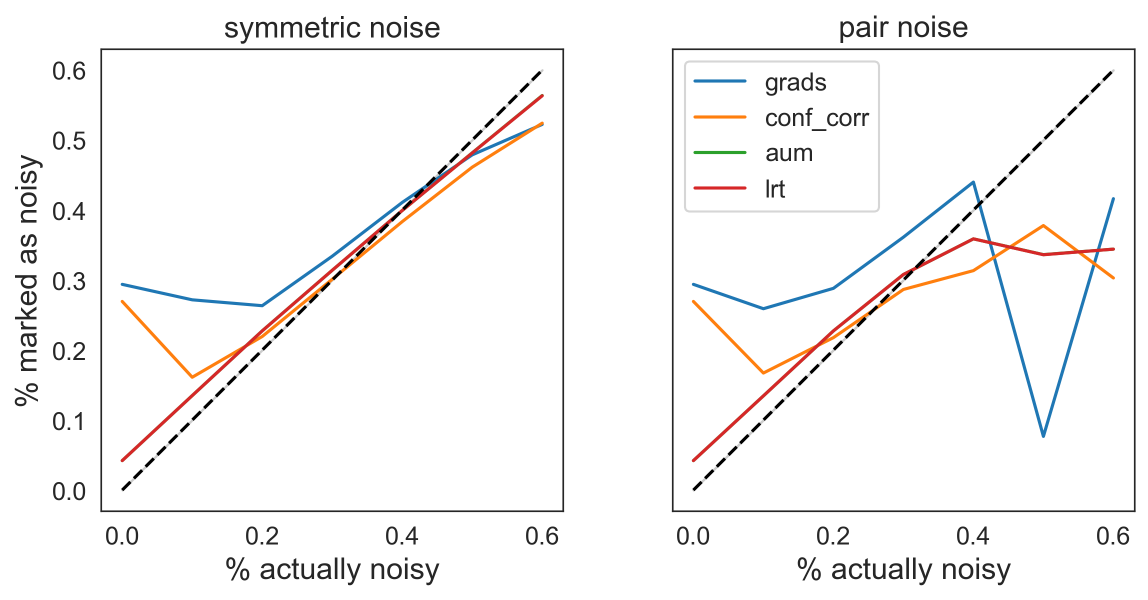}
\label{fig:infer_noise_rate_all_bean}
\end{figure}

Figure \ref{fig:infer_noise_rate_all_bean} maps the actual noise rate against the percentage of instances marked as noisy by the noise detection methods at the first epoch after warm-up. The black dotted line represents an ideal noise detection classifier, where the percentage of instances marked as noisy matches the actual noise rate present in the data.\textit{LRT} and \textit{AUM} overlap, which is why \textit{AUM}'s line isn't visible. 

\textit{LRT} and \textit{AUM} generally mark a number of instances as noisy that is close to the actual amount present in the data. At the lower noise rates, both \textit{Gradients} and \textit{ConfCorr} tend to label more instances as noisy than are actually present in the datasets. This could indicate that they are more sensitive to outliers or class imbalances. 
Consequently, the percentage of instances marked as noisy by \textit{LRT} and \textit{AUM} can be used to approximate the level of noise present in datasets polluted by symmetric noise. For pair noise exceeding 40\%, these estimates become increasingly unreliable. 

\begin{figure}[H]
\caption{Metrics calculated at different epochs with 30\% pair noise (Dry Bean). The ability to discriminate between clean and noisy instances remains roughly constant during training for most methods. }
\centering
\includegraphics[width=0.89\textwidth]{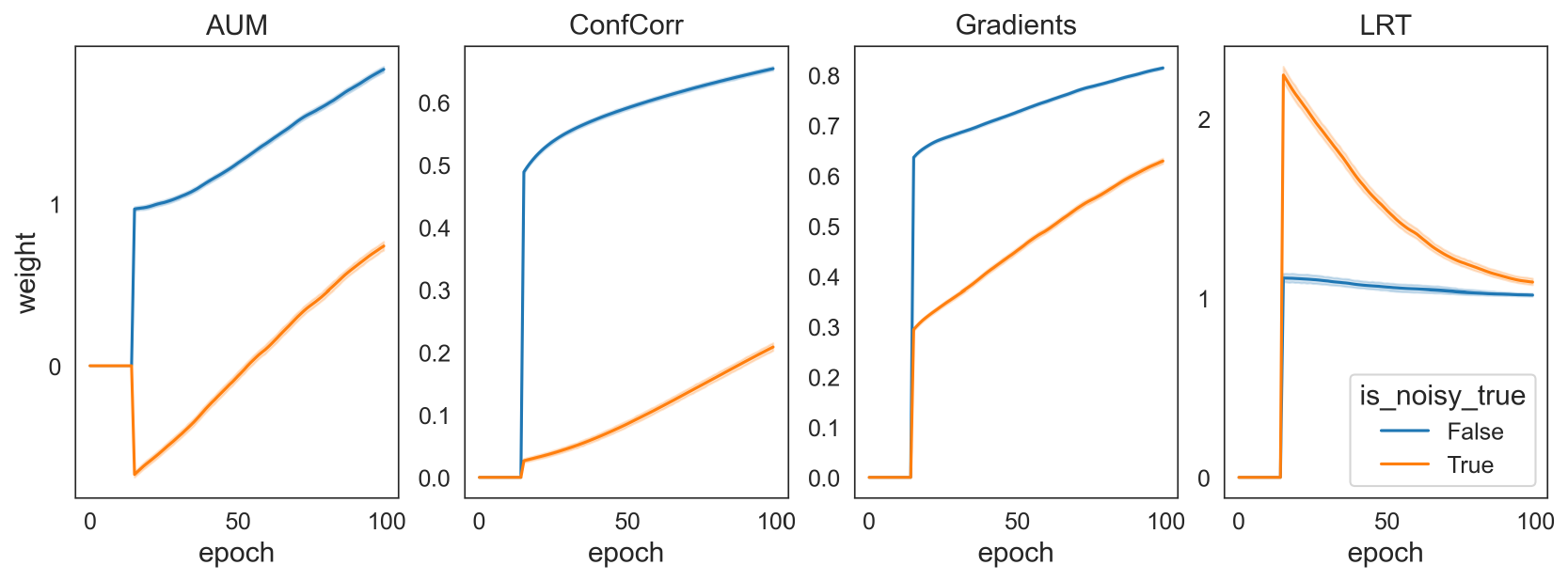}
\label{fig:pair_crit_epoch}
\end{figure}

During training, \textit{AUM},  \textit{ConfCorr} and \textit{Gradients} all increase (Figure \ref{fig:pair_crit_epoch}), but the distance between noisy and clean instances stays roughly the same. \textit{LRT}s metrics decrease and converge as training progresses.

This means that metrics for noisy and clean instances tend to converge as training progresses. Depending on the metric, both values either increase or decrease. The values calculated on noisy instances tend to  have a steeper slope, meaning that they change more drastically during training than those calculated on clean instances. 

The ability of all methods to discriminate between clean and noisy instances remains roughly constant during training, with the exception of \textit{LRT}. The convergence for \textit{LRT}'s plot shows that the model adapts to noisy instances as training progresses and therefore noisy and clean instances become increasingly harder to differentiate, highlighting again that early stopping is an important tool to avoid overfitting noisy labels. For all methods, using a fixed threshold without early stopping will likely be ineffective in discriminating between clean and noise instances throughout the whole training procedure.


\begin{figure}[H]
\caption{Logloss and accuracy on the test set per epoch when removing or relabeling instances marked as noisy by a noise detection method, calculated on 30\% pair noise (Covertype).}
\centering
\includegraphics[width=0.7\textwidth]{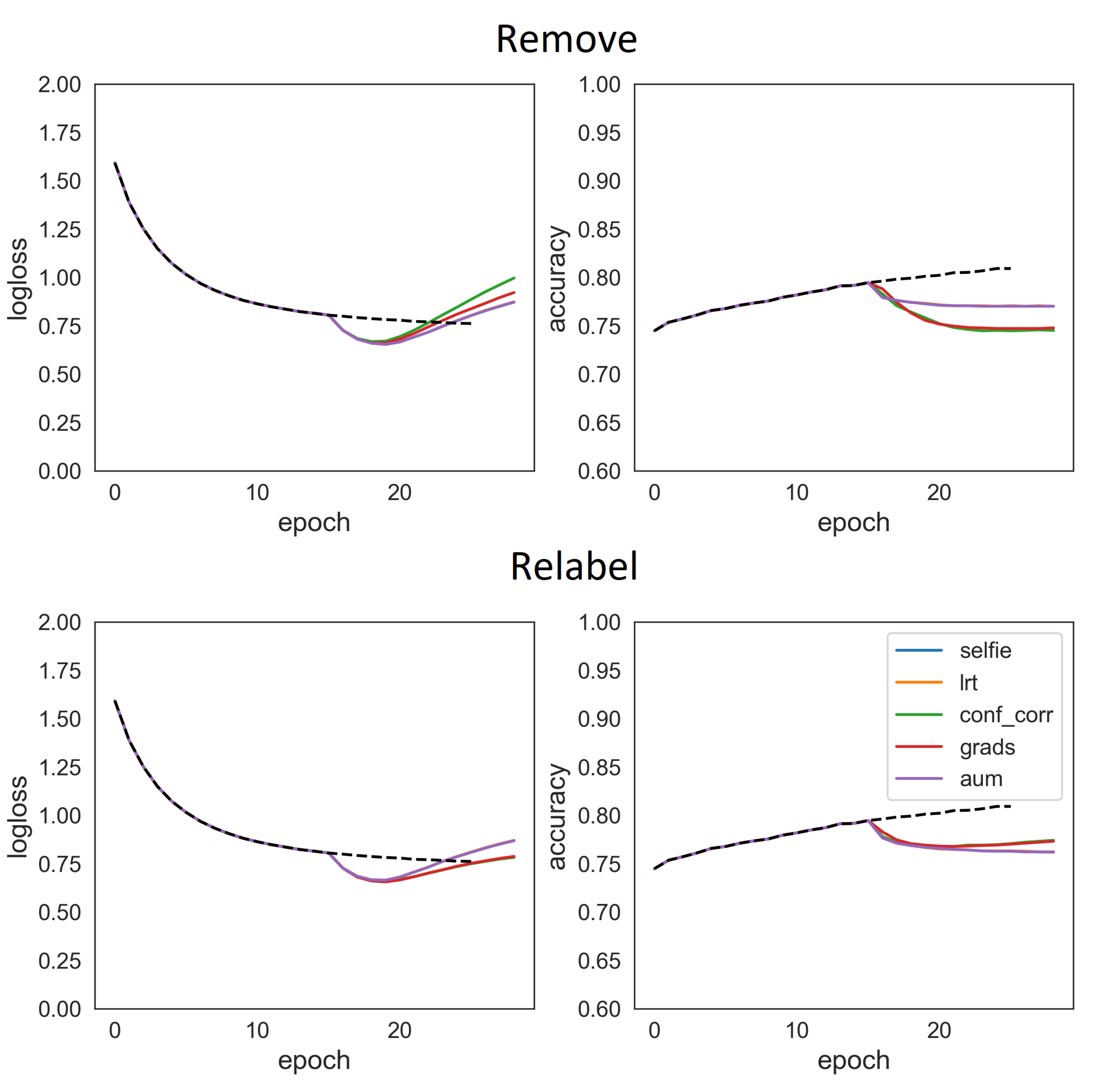}
\label{fig:remove_acc_error}
\end{figure}

Figure \ref{fig:remove_acc_error} visualizes the training curves when removing or relabeling noisy candidate instances detected using the corresponding detection method. The black dotted line represents the baseline model, i.e. not correcting for label noise, and each detection method is represented by a colored line. For the duration of the warm-up period, which lasts 15 epochs, all methods perform the same. After that there is a significant drop in error for all detection methods, while the baseline model's error stays high. But at the same time, accuracy decreases for all models. 

With regard to the drop in error, it can be concluded that early stopping alone is not sufficient to robustly train a GBDT model on data containing noisy labels. Instead, removing or relabeling suspicious instances results in a significant drop in error. But the decrease in loss did not translate to an increase in accuracy. Though \textit{AUM} combined with removal achieves a higher accuracy than the other methods, it's still worse than the baseline model. In Section \ref{sec:final-results} we'll look at the classification accuracy in diverse noise settings and different datasets to find out whether noise treatment actually worsens classification performance. 

Removal and relabeling appear to perform roughly the same according to Figure \ref{fig:remove_acc_error}.  The drop in accuracy is slightly lower with relabeling as well as the increase in error after about twenty epochs. Based solely on this information, a definitive conclusion regarding the best correction method cannot be determined. 

\begin{figure}[H]
\caption{The types of instances that got marked as noisy by \textit{AUM} per epoch, calculated at 30\% pair noise (Covertype). About 40\% of the instances marked as noisy are actually clean.}
\centering
\includegraphics[width=0.6\textwidth]{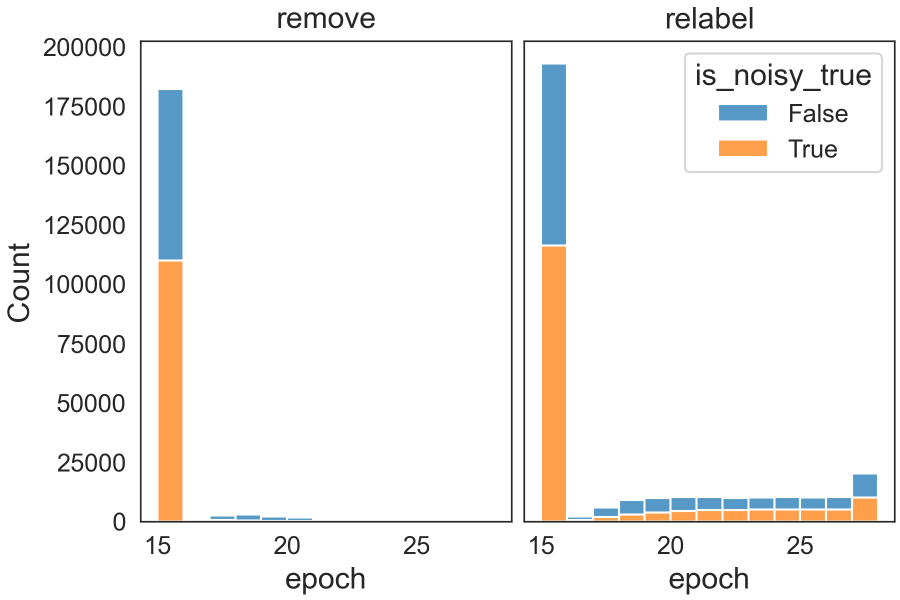}
\label{fig:type_instance_remove_relabel}
\end{figure}

As an example, Figure \ref{fig:type_instance_remove_relabel} shows which instances were marked as noisy by \textit{AUM} at each epoch. At the first epoch after warm-up, about  58\% of the instances marked as noisy were in fact noisy. Since these instances are removed, resulting in a reduced training set size, only a few instances are marked as noisy in subsequent epochs, most of which are actually clean. When relabeling noisy candidates, a small number of instances are selected for relabeling each epoch, with about half of them being clean. However, even though no instances are removed during relabeling, each instance can only be relabeled once. This implies that, when using relabeling, more instances are identified as noisy, possibly due to the relabeled instances influencing the model's perception of the remaining instances. On one hand, relabeling is beneficial as it utilizes the entire dataset, but on the other hand, it could introduce more errors and thus potentially degrade model performance. Regarding the instances marked as noisy by the other detection methods in the first epoch after warm-up, about 60\% are actually noisy. The same goes for all the methods when trained on 30\% symmetric noise (see online code). 


\begin{figure}[H]
\caption{Logloss per quantile of the top noise values to be treated. (Dry Bean). Treating fewer instances than the actual percentage of noisy instances tends to yield the best performance.}
\centering
\includegraphics[width=0.79\textwidth]{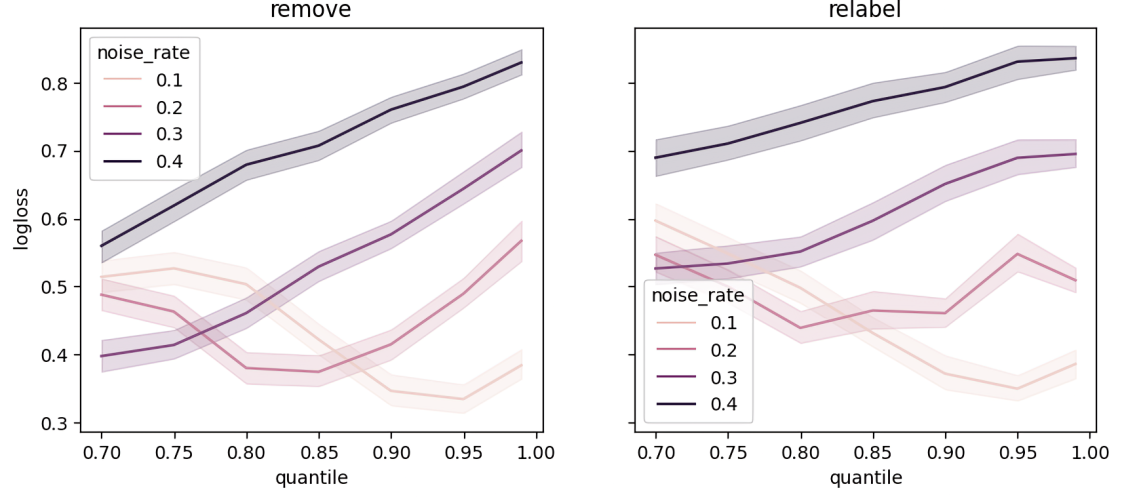}
\label{fig:loss-quantile}
\end{figure}

Instead of setting a fixed upper limit for the percentage of instances that can be removed, another approach is to define a threshold based on the top quantile of noise values, with instances above this threshold being marked for treatment. For example, when the threshold is set at the 90th percentile, the 10\% of instances with the highest noise values are selected for treatment. Figure \ref{fig:loss-quantile} shows the log-loss on the test set across different quantiles for various noise rates. The minima of the curves are relatively equidistant, suggesting that treating fewer instances than the actual proportion of noisy instances typically yields the best performance. This aligns with earlier findings showing that over one-third of the instances marked as noisy are, in fact, clean. Therefore, in cases where the level of noise can be estimated, adjusting the proportion of instances treated can lead to better outcomes. This method, however, is not employed in the results section and represents an avenue for further research. 

\subsection{Comparison with the State-of-the-Arts}\label{sec:final-results}
This section numerically presents the noise detection and classification performance results in various noise settings and on different datasets. The performance of the implemented methods is also compared to state-of-the-art research using the same metrics, datasets and similar noise settings. The following tables were generated at the most optimal epoch as indicated by early stopping. In them, the best values per category are marked in bold letters.  

In their paper, \cite{training-dynamics} implemented a variant of pair noise.  On the Adult dataset, they achieved a maximum noise detection accuracy of 86.93\%, 83.97\% and 82.58\% at noise rates from 10\% to 30\% each. On the same dataset, the methods implemented in this work achieve a maximum noise detection accuracy of 99.70\%, 99.20\% and 98.56\% at noise rates from 10\% to 30\% each (Table \ref{tab:detection_accuracy_adult}), thereby outperforming Ponti et al.'s work across all noise rates. But none of the implemented methods managed to outperform their work on the Covertype and Breast Cancer datasets. 




In this work, \textit{AUM} and \textit{LRT} performed best on the Adult dataset across all noise rates (Table \ref{tab:detection_accuracy_adult}), whereas on the Breast Cancer dataset, \textit{ConfCorr} and \textit{Gradients} are best at detecting noise (Table \ref{tab:detection_accuracy_cancer}). On the Covertype dataset, \textit{ConfCorr} reached the highest noise detection performance across all metrics. These findings demonstrate that the optimal noise detection method varies depending on the dataset it is applied to.

\begin{table*}[h!]
\centering
\begin{minipage}{0.48\textwidth}
\centering
\begin{tabular}{@{\hskip 0pt}rlr@{\hskip 0pt}}
\hline
Noise Rate & Detection & Accuracy \\
\hline
0.10 & AUM & \textbf{99.70} \\
 & ConfCorr & 93.30 \\
 & Gradients & 94.00 \\
 & LRT & \textbf{99.70} \\
0.20 & AUM & \textbf{99.20} \\
 & ConfCorr & 94.40 \\
 & Gradients & 98.00 \\
 & LRT & \textbf{99.20} \\
0.30 & AUM & 98.50 \\
 & ConfCorr & 98.44 \\
 & Gradients & \textbf{98.56} \\
 & LRT & 98.50 \\
\hline
\end{tabular}
\caption{Noise detection performance on the Adult dataset with 10\% to 30\% pair noise. Highest values are marked by noise rate.}
\label{tab:detection_accuracy_adult}
\end{minipage}%
\hfill
\begin{minipage}{0.48\textwidth}
\centering
\begin{tabular}{@{\hskip 0pt}rlr@{\hskip 0pt}}
\hline
Noise Rate & Detection & Accuracy \\
\hline
0.10 & AUM & 80.94 \\
 & ConfCorr & \textbf{83.06} \\
 & Gradients & 77.20 \\
 & LRT & 80.94 \\
0.20 & AUM & 80.90 \\
 & ConfCorr & \textbf{82.90} \\
 & Gradients & 79.90 \\
 & LRT & 80.90 \\
0.30 & AUM & 79.60 \\
 & ConfCorr & \textbf{81.00} \\
 & Gradients & 78.80 \\
 & LRT & 79.60 \\
\hline
\end{tabular}
\caption{Noise detection performance on the Covertype dataset with 10\% to 30\% pair noise. Highest values are marked by noise rate.}
\label{tab:detection_accuracy_cover}
\end{minipage}
\end{table*}

\begin{table}[h!] 
\centering
\begin{tabular}{@{\hskip 0pt}rlr@{\hskip 0pt}}
\hline
Noise Rate & Detection & Accuracy \\
\hline
0.10 & AUM & 90.30 \\
 & ConfCorr & \textbf{92.06} \\
 & Gradients & 90.56 \\
 & LRT & 90.30 \\
0.20 & AUM & 80.44 \\
 & ConfCorr & \textbf{90.75} \\
 & Gradients & \textbf{90.75} \\
 & LRT & 80.44 \\
0.30 & AUM & 72.06 \\
 & ConfCorr & \textbf{85.25} \\
 & Gradients & 82.60 \\
 & LRT & 72.06 \\
\hline
\end{tabular}
\caption{Noise detection performance on the Breast Cancer dataset with 10\% to 30\% pair noise. Highest values are marked by noise rate. }
\label{tab:detection_accuracy_cancer}
\end{table}

It's unclear whether the performance of the final classification results is calculated on a specific noise rate or averaged between the tested noise rates of 10\% to 30\% in Ponti et al.'s paper. If the former is true, it would skew their result in a more positive direction. On the Adult dataset, Ponti et al. achieved a maximum precision, recall and f1-score of 0.58, 0.89 and 0.70 each. Therefore this work achieved a higher precision (79.56\%) than Ponti et al. but a lower recall (57.03\%) and f1-score (66.06\%)(Table \ref{tab:clf_performance_adult}). 

On the Covertype dataset, Ponti et al. achieved a maximum precision, recall and f1-score of 0.78, 0.74 and 0.75 respectively. This work achieved a higher precision (78.80\%) than Ponti et al. but a lower recall (62.44\%) and f1-score (66.06\%) (Table \ref{tab:clf_performance_cover}).

On the Breast Cancer dataset, Ponti et al. achieved a maximum precision, recall and f1-score of 1.00, 0.86 and 0.92 respectively. This work achieved a lower precision (94.10\%) than Ponti et al. but a higher recall (90.25\%) and about the same f1-score (91.56\%) (Table \ref{tab:clf_performance_cancer}) using \textit{ConfCorr} or \textit{Gradients} in combination with noisy candidate removal. 

\textit{ConfCorr} and \textit{Gradients} performed best in terms of classification accuracy on the Adult dataset with 30\% noise, especially when paired with relabeling as the noise correction method. On the Breast Cancer dataset, those methods performed best when combined with noisy candidate removal instead of relabeling. In the case of the Covertype datasets, attempting to correct for noise seemed to be counterproductive, potentially introducing more errors or eliminating informative instances. Not correcting for noise resulted in the highest classification performance in terms of accuracy, recall, and f1-score. The sole exception was \textit{ConfCorr} combined with removal, which attained the highest precision score. This shows that the ideal noise correction method is dependent on both the dataset it is applied to and the noise detection method it is paired with.


\begin{table}[h!] 
\centering
\begin{tabular}{@{\hskip 0pt}l@{\hskip 6pt}l@{\hskip 6pt}l@{\hskip 6pt}l@{\hskip 6pt}l@{\hskip 6pt}l@{\hskip 0pt}}
\hline
Correction & Detection & Accuracy & Precision & Recall & F1 \\
\hline
none & none & 85.94 & 77.60 & 56.94 & 65.70 \\
relabel & AUM & 86.00 & 78.40 & 56.22 & 65.50 \\
 & ConfCorr &\textbf{86.10} & 78.44 & \textbf{57.03} & \textbf{66.06} \\
 & Gradients & \textbf{86.10} & 78.44 & \textbf{57.03} & 66.00 \\
 & LRT & 86.00 & 78.40 & 56.22 & 65.50 \\
remove & AUM & 86.00 & 78.44 & 56.20 & 65.50 \\
 & ConfCorr & 85.60 & \textbf{79.56} & 52.78 & 63.47 \\
 & Gradients & 85.50 & 78.75 & 52.70 & 63.16 \\
 & LRT & 86.00 & 78.40 & 56.16 & 65.44 \\
\hline
\end{tabular}
\caption{Classification performance on the Adult dataset with 30\% pair noise.  } 
\label{tab:clf_performance_adult}
\end{table}

\begin{table*}[t] 
\centering
\begin{tabular}{llrrrr}
\hline
Correction & Detection & Accuracy & Precision & Recall & F1 \\
\hline
none & none & \textbf{76.94} & 78.06 & \textbf{62.44} & \textbf{66.06} \\
relabel & AUM & 73.75 & 76.56 & 56.16 & 59.44 \\
 & ConfCorr & 74.60 & 76.75 & 57.34 & 61.16 \\
 & Gradients & 74.50 & 76.60 & 57.16 & 60.72 \\
 & LRT & 73.70 & 76.30 & 56.20 & 59.44 \\
remove & AUM & 74.44 & 78.00 & 58.70 & 62.90 \\
 & ConfCorr & 72.60 & 77.06 & 52.97 & 53.30 \\
 & Gradients & 73.30 & \textbf{78.80} & 52.12 & 53.06 \\
 & LRT & 74.44 & 78.00 & 58.60 & 62.75 \\
\hline
\end{tabular}
\caption{Classification performance on the Covertype dataset with 30\% pair noise. } 
\label{tab:clf_performance_cover}
\end{table*}

\begin{table*}[t] 
\centering
\begin{tabular}{llrrrr}
\hline
Correction & Detection & Accuracy & Precision & Recall & F1 \\
\hline
 none & none & 90.40 & 96.94 & 87.50 & 92.00 \\
    relabel & AUM & 91.25 & 94.30 & \textbf{91.70} & 92.94 \\
     & ConfCorr & 90.40 & 95.50 & 88.90 & 92.06 \\
     & Gradients &  90.40 &  95.50 &  88.90 &  92.06 \\
     & LRT &  90.40 &  95.50 &  88.90 &  92.06 \\
    remove & AUM & 91.25 & 94.30 & \textbf{91.70} & 92.94 \\
     & ConfCorr & 92.10 & 97.00 & 90.25 & 93.50 \\
     & Gradients & \textbf{93.00 }& \textbf{97.06} & \textbf{91.70} & \textbf{94.30} \\
   & LRT & 91.25 & 95.56 & 90.25 & 92.90 \\
\hline
\end{tabular}

\caption{Classification performance on the Breast Cancer dataset with 30\% pair noise. } 
\label{tab:clf_performance_cancer}
\end{table*}

\subsection{Summary}

Our results demonstrate that GBDTs show a natural robustness to label noise, particularly to symmetric noise. By implementing early stopping, we effectively prevent GBDTs from adapting to noisy labels, thereby preserving test accuracy.

Both \textit{AUM} and \textit{LRT} achieved state-of-the-art noise detection accuracy, reaching 99\% across all noise rates on the Adult dataset, and proved effective for estimating the amount of noise present in the data. In addition, \textit{ConfCorr} demonstrated high classification precision on the Adult dataset, while \textit{Gradients} achieved superior classification precision and recall on the Covertype and Breast Cancer datasets, respectively, outperforming other approaches.

The use of dynamic thresholding with a Gaussian Mixture Model (GMM) for \textit{ConfCorr} and \textit{Gradients} further enhanced performance by eliminating the need for additional parameter tuning, allowing these methods to adapt to varying noise levels automatically.

\section{Discussion} \label{sec:discussion}
Our findings suggest that noise detection performance is highly dependent on dataset characteristics, noise type, and noise rate, indicating that the optimal combination of detection and correction methods is context-specific. Class imbalance also presents challenges, as some methods may inadvertently remove instances from smaller classes, potentially reducing model performance. 

Relabeling noisy instances can sometimes introduce additional errors, while removing them might lead to the loss of informative data. Consequently, multiple methods should be evaluated to identify the best approach for managing label noise in any given scenario. A cautious approach is recommended, generally treating fewer instances than the estimated noise level to mitigate potential drawbacks.

All methods exhibited sensitivity to class imbalance, with performance notably lower on the highly imbalanced Covertype dataset. Furthermore, time and computational constraints limited the study to a single trial per experiment. The observed variability in method performance across datasets suggests that favorable outcomes may occasionally result from chance rather than inherent effectiveness, underscoring the value of cross-validation for greater reliability. 

Finally, the use of simple noise types, such as pair and symmetric noise, may limit the real-world applicability of these results, as they do not fully represent the complexity of real-world label noise.

\section{Conclusion}\label{sec:conclusion}
This work examines the impact of label noise on Gradient Boosted Decision Trees (GBDTs) and evaluates several label noise detection and correction methods tailored for tabular data—an underexplored area in label noise research. We adapted two noise detection techniques from deep learning for use with GBDTs and further developed two methods specifically for this study: the \textit{Gradients} method, inspired by weight clipping in Adaboost and the small loss trick, and an extension of \textit{ConfCorr} to support relabeling.

Our experiments demonstrate that GBDTs exhibit natural robustness to label noise in the early training stages, particularly in cases of symmetric noise, where early stopping proved effective in preserving test accuracy. Both \textit{AUM} and \textit{LRT} achieved state-of-the-art noise detection accuracy on the Adult dataset, with the percentage of instances marked as noisy by these methods providing a useful estimate of the dataset’s noise level. In addition, \textit{ConfCorr} achieved high classification precision on the Adult dataset, while \textit{Gradients} outperformed other methods in terms of classification precision and recall on the Covertype and Breast Cancer datasets, respectively.

The inconsistency in method performance across different datasets indicates that optimal noise detection and correction strategies may depend on factors such as dataset characteristics, noise rate, and noise type. This variability suggests that favorable results may, in some cases, stem from dataset-specific factors rather than inherent method effectiveness. Repeated experiments and cross-validation could provide greater clarity and reliability in assessing method performance.

Future research could explore alternative relabeling strategies beyond relying on the most confident or frequent predictions from recent epochs. Since high noise detection accuracy does not always lead to improved classification performance, further corrective measures, such as error correction, may enhance the utility of noise detection methods. Additionally, given that classification performance often decreased when removing or relabeling noisy instances in imbalanced datasets, future studies could develop noise detection techniques better suited to class-imbalanced data in GBDTs.

As GBDTs continue to generally outperform Deep Neural Networks (DNNs) on tabular data, this work contributes essential insights into label noise handling for these models. Should DNNs eventually surpass GBDTs in tabular tasks, it would be valuable to investigate the applicability of noise handling techniques from image and text domains for tabular data. This study serves as a foundational step in advancing robust GBDT training under noisy conditions, laying the groundwork for continued exploration in this critical area of machine learning.

\section{CRediT Authorship Contribution Statement and AI Disclosure}
Generative AI tools, such as ChatGPT, were used under human oversight solely to improve the readability and language of this manuscript. No generative AI was used for data analysis, content generation, or any other aspect of the research process. 
\textbf{Anita Eisenbürger: } Conceptualization, Data Curation, Formal analysis, Investigation, Methodology, Visualization, Writing -- original draft, Writing -- review \& editing. \textbf{Daniel Otten: } Supervision, Writing -- review \& editing. \textbf{Frank Hopfgartner: } Supervision, Writing -- review \& editing. \textbf{Anselm Hudde: } Writing -- review \& editing. 

\bibliographystyle{elsarticle-harv} 
\bibliography{main}

\end{document}